# Saliency Revisited:
Analysis of Mouse Movements versus Fixations

Hamed R. Tavakoli, Fawad Ahmed,
Ali Borji, Jorma Laaksonen

# Saliency Revisited: Analysis of Mouse Movements versus Fixations

Hamed R. Tavakoli[†]    Fawad Ahmed[‡]    Ali Borji[‡]    Jorma Laaksonen[†]
[†]Dept. of Computer Science, Aalto University, Finland
[‡]Dept. of Computer Science, University of Central Florida, Orlando

## Abstract

*This paper revisits visual saliency prediction by evaluating the recent advancements in this field such as crowd-sourced mouse tracking-based databases and contextual annotations. We pursue a critical and quantitative approach towards some of the new challenges including the quality of mouse tracking versus eye tracking for model training and evaluation. We extend quantitative evaluation of models in order to incorporate contextual information by proposing an evaluation methodology that allows accounting for contextual factors such as text, faces, and object attributes. The proposed contextual evaluation scheme facilitates detailed analysis of models and helps identify their pros and cons. Through several experiments, we find that (1) mouse tracking data has lower inter-participant visual congruency and higher dispersion, compared to the eye tracking data, (2) mouse tracking data does not totally agree with eye tracking in general and in terms of different contextual regions in specific, and (3) mouse tracking data leads to acceptable results in training current existing models, and (4) mouse tracking data is less reliable for model selection and evaluation. The contextual evaluation also reveals that, among the studied models, there is no single model that performs best on all the tested annotations.*

## 1. Introduction

There has been a significant recent progress in the field of visual saliency. Numerous models and datasets have been introduced. The new databases have been expanded along two dimensions: (1) increasing the number of images and viewers, and (2) introducing richer contextual annotations (e.g., image categories [2], and regional attributes [33], etc.). To accomplish these objectives, researchers have been relying on crowd sourcing schemes for recording eye movements (e.g., using webcams [34]) or alternative signals such as mouse movements and clicks (e.g. [14]), and annotations. Along with these advances, however, new challenges have surfaced that need to be addressed. For example, it remains to be answered whether and to what degree different at-tentional proxies agree with each other? Is it possible to reach human level accuracy by utilizing large scale mouse data? and how should these new types of data be used for saliency model evaluation and construction? Fig. 1 visually compares density maps from eye and mouse tracking. It depicts a noticeable difference between maps encouraging a detailed quantitative study.

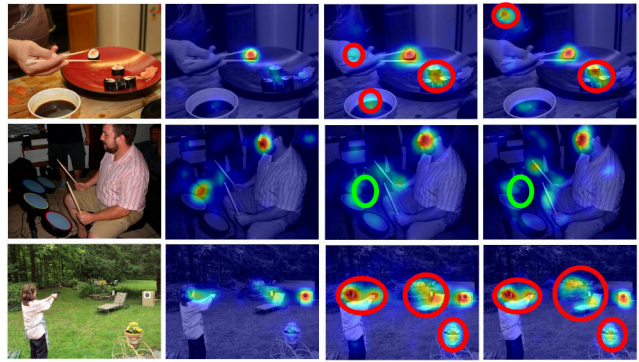

Figure 1. Visual comparison of fixation maps using eye and mouse tracking, overlaid on images. From left to right: image, eye tracking, mouse tracking using Amazon Mechanical Turk (AMT), and mouse tracking in controlled laboratory (LAB) from [33]. The red and green ellipses indicate over and under estimation, respectively.

**Our contributions.** This paper presents 2 main contributions: (1) assessing the quality of crowd sourced mouse tracking as an alternative to eye tracking and the effect of such data on model training and evaluation, and (2) introducing a contextual evaluation scheme for evaluating models in a fine-grained manner. The contextual evaluation is applicable to both model assessment and comparison of mouse tracking with eye tracking. Further, this study will be addressing some of the questions that have surfaced by introduction of mouse tracking based databases and help better understand saliency models' performance.

**Is this another benchmark?** No. We are not carrying out yet another benchmark. The literature is already replete with benchmarks [3, 17, 4, 6], metric discussions [21, 26, 32, 7], review papers [28, 1], and analytical model comparisons [8]. Alternatively, we seek to answer some important questions pertaining to mouse tracking as a substitute to eye movements. The answers will benefit both model construction and evaluation. We furthermore attempt to incorporate available contextual annotations into model evaluation in order to facilitate automatic detailed analysis of models. This will help us understand the strengths and weaknesses of models better.

## 2. Related studies

Considering mouse tracking analysis for visual saliency and fixation prediction, the most relevant work is [14]. It proposes to use mouse tracking instead of eye tracking in order to scale up data collection over stimuli of larger magnitude (millions instead of hundreds or thousands). Jiang et al. [14] analyzed some properties of mouse data such as center-bias, evaluated mouse maps against fixation maps, and compared saliency models using mouse tracking data. There are, however, some aspects which are overlooked in their work such as congruency among participants, the effect of mouse tracking on training a saliency model, model evaluation and comparison using mouse versus eye tracking. Here, we revisit Jiang et al.'s work and conduct a systematic investigation of mouse data with respect to eye movements.

Considering contextual model analysis, the most relevant to this paper is the recent study by Bylinskii et al. [8]. They chose several top-performing saliency models based on deep learning architectures over the MIT300 dataset [6]. They then conducted a behavioral study by asking Amazon Mechanical Turk workers to label (1 out of 15 labels) image regions falling on $> 95\%$ of the fixation heatmap. Analyzing failure of models on those regions, they found that about half the errors made by models are due to failures in accurately detecting parts of people, faces, animals, and text which means that models should try to improve on those areas. The main criticism to such a human-based evaluation is that it is limited to the number of models, subjects, and images making it expensive to conduct in large scale. To address such shortcomings, we extend the existing evaluation schemes and propose a systematic framework for contextual model evaluation. We then employ the proposed technique for detailed comparison of mouse tracking and eye tracking whenever possible.

The benchmark and metric studies have some commonalities with the current study in terms of methodology. For example, Borji et al. [4] analyzed different parameters affecting saliency evaluation (e.g., center-bias, class categories, etc) in order to benchmark models. Riche et al. [26] employed statistical analysis in order to do metric selection for saliency evaluation. They show a small number of metrics is enough for model evaluation as many metrics carry similar information. This study is, however, addressing the impact of mouse tracking as an alternative to eye tracking for learning saliency.

| Database | Tracking Technology | | | (O, I) | Contextual Annotation | | | |
|---|---|---|---|---|---|---|---|---|
| | ET | wET | MT | | OL | IC | OA | FB |
| FIGRIM [5] | × | — | — | (15,630) | × | × | — | — |
| EFC [15] | × | — | — | (16,500) | × | — | × | — |
| KTH Koostra [19] | × | — | — | (31, 99) | — | × | — | — |
| NUSEF [25] | × | — | — | (25, 758) | — | × | — | — |
| CAT2000 [2] | × | — | — | (24, 4000) | — | × | — | — |
| salObj [22] | × | — | — | (12, 850) | × | — | — | — |
| iSUN [34] | — | × | — | (3, 8926) | × | × | — | — |
| SALICON [14] | — | — | × | (60, 10000) | × | — | — | — |
| OSIE [33] | × | — | × | (15, 700) | × | — | × | × |

Table 1. Comparison of databases in terms of augmented annotations and eye tracking technology. (O, I) corresponds to the average number of observers and the number of images. The tracking technologies are ET: commercial high-end eye tracking devices, wET: webcam based eye tracking, MT: mouse tracking. Type of contextual information are OL: object type and localization, IC: image category and scene type, OA: object attributes, FB: foreground/background property of objects.

## 3. Saliency databases & contextual annotation

There exists numerous databases for saliency evaluation. As of this writing, 23 datasets are enumerated by [6]. These databases are often compared with each other in terms of stimuli and experimental setup (e.g., number of observers, distance to image center, recording device, task, etc.). Instead of comparing datasets along these dimensions, we study them in terms of augmentation with extra information. There are different levels of augmentation including, image class categories, object localization, and object attribute annotations.

Table 1 summarizes the information of some of the most notable augmented databases. As depicted, most of the databases have object category annotations in terms of object bounding boxes, object masks or object boundaries. There are, however, only two databases with contextual object attribute annotations: Eye Fixation in Crowd (EFC) [15] and Object and Semantic Images and Eye-tracking (OSIE) [33]. The EFC database is collected for analyzing saliency in crowds. It contains face bounding box localizations and their attributes such as if a face is frontal, profile, back or occluded.

The OSIE database has the widest range of contextual annotations consisting of twelve boolean attributes conveying semantic meaning of objects. These attributes include: Text, Face (includes: back, profile, and frontal faces), Emotion (if a face conveys emotion), Sound (objects producing sound), Smell (objects with a scent), Taste (anything that can be tasted), Touch (anything with tactile feeling), Motion (moving/flying object), Operability (natural or man-made tools used by holding with hands), Watchable (man-made objects designed to be watched), Touched (an object being touched), and Gazed (if an object is gazed by someone in the image). Besides rich contextual annotations, OSIE defines the foreground/background property of objects, which is desirable to validate how well a model discriminates background and foreground regions.

The OSIE dataset provides an invaluable opportunity for comparing mouse and eye tracking. It consists of 700 images, 15 observers for eye tracking (OSIE EYE), approximately 46 mouse tracking participants using a controlled laboratory setup (OSIE LAB), and at least 86 mouse tracking participants using Amazon Mechanical Turk (OSIE AMT). Thus, we base our study on OSIE dataset.

## 4. Metrics

**Metrics of saliency evaluation.** Numerous metrics for saliency evaluation have been introduced in the past. Some of them are: ROC-based metrics (e.g., AUC [30], AUC-Judd [18], AUC-Borji [4], shuffled AUC (sAUC) [4, 36], binned AUC [32]), similarity-based metrics (e.g., Correlation Coefficient (CC) [16], Kullback-Leibler divergence (KL) [21, 36, 13], Similarity score (SIM) [17], Earth mover's distance (EMD) [17], Information Gain (IG) [20]), Normalized scanpath saliency (NSS) [24], and metrics based on fixation sequence (e.g., Scanpath evaluation score [4]). For brevity, we skip explaining them in details and refer the readers to relevant publications [4, 7, 21, 26].

Minor variations in metrics can sometimes have significant consequences in the metric interpretation. For example, [13] employ KL metric as a technique to measure the similarity between the distribution of fixated and random locations in a saliency map, while in [6, 26], the KL is measured in terms of the similarity between fixation density maps and saliency maps. Consequently, in [13] a higher KL value is better and in [6] a lower value is superior. It is worth noting that we follow [6] in our experiments.

**The appropriate metrics.** Many of the saliency metrics convey the same information making model performance interpretation difficult. For easier interpretability of the results, we are motivated to select a subset of metrics. To this end, mouse tracking data, OSIE AMT, is evaluated against human eye fixation data, OSIE EYE. For the metrics, the Spearman's rank correlation coefficient ($\rho$) is computed between the score values of images. Classical multidimensional scaling (MDS) is then employed for 2D visualization of the correlation matrix. Results are summarized in Fig. 2, indicating an overall high correlation between the metrics. Considering the projection on the first eigenvalue ($x$-axis) — it is the most contributing eigenvalue —, metrics can be grouped into three clusters. The biggest cluster includes metrics that encode fixation information, AUC-based metrics as well as NSS and IG. The other two clusters consist of (1) CC and SIM, and (2) EMD and KL. We choose **SIM**, **KL**, and **sAUC** for reporting the performance of models. The sAUC is preferred over other AUC metrics and NSS because (a) it has well-defined lower and upper bound values, (b) it has a defined chance-level value, and (c) it accounts for center-bias in fixation distributions [4]. SIM and KL metrics are selected as they act complementary to each other, according to Fig. 2.

**Algorithm 1** Computing metrics of contextual evaluation for a saliency map: how to scale your preferred conventional metric for exploiting contextual data. $\odot$ is the element-wise product.

**Input:** $Sal$ : a saliency map of size $W \times H$, a tensor of contextual masks $Cm$ of size $W \times H \times O$, where $O$ is the number of regions, the contextual attribute matrix $Ca$ of size $O \times N$, which reports existence of an attribute, where $N$ is the number of attributes, and human fixation map $Fix$ of size $W \times H$.
**Output:** A vector of contextual attributes' evaluation $Score$ of size $N$.
1: **for all** $o$ regions in $Cm$ **do**
2:     $Sal_o = Sal \odot Cm(:,:,o)$
3:     $Fix_o = Fix \odot Cm(:,:,o)$
4:     s = compute_metric($Sal_o$, $Fix_o$)
5:     **for all** $n$ attributes in $Ca$ **do**
6:        **if** $Ca(o,n)$ is true **then**
7:           update_mean($Score(n)$, s)
8:        **end if**
9:     **end for**
10: **end for**

**Contextual saliency evaluation.** To perform contextual evaluation, we use existing metrics with regard to contextual annotations. That is, given an image, we employ the existing metrics within specified regions of images, which are associated with contextual attributes. Algorithm 1 presents how to compute the agreement between human eye fixations and saliency maps inside annotated regions associated with attributes such as gaze, face, and text.

In principle, all existing saliency evaluation metrics can be employed for the purpose of contextual evaluation by the proposed algorithm. The result of such a contextual evaluation is a vector of scores that helps investigating the pros and cons of a model capturing each property. While we recommend using the contextual scores for a fine-grained analysis, it is also possible to summarize the scores into one for the purpose of model ranking. To achieve this, given the score vector of a saliency map, $Score$, we define a weighted average score as $CScore = \sum_{n=1}^{N} w_n Score_n$ where $N$ is the number of contextual attributes and $w$ is the weight vector indicating the importance of each property. $\sum_{n=1}^{N} w_n = 1$, where:

$$w_n = \frac{\text{\# of fixations on attribute } n}{\text{\# of fixations on images with attribute } n}. \quad (1)$$

To further summarize the scores over a database, the average over scores, *mean CScore*, is employed.

It is worth noting that all the attributes may not be present in all the images when computing the average contextual score. Thus, the average should be done with respect to the number of images having an attribute, and not all images.

## 5. Analysis of mouse tracking data

**Inter-participant visual congruency.** *"How is the inter-participant visual congruency (IPVC) on mouse tracking*

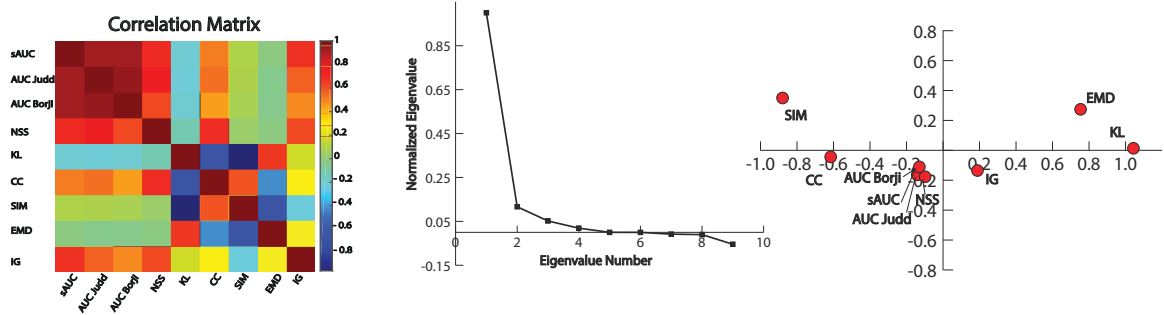

Figure 2. Selecting the appropriate metric using multidimensional scaling (MDS) analysis. From left to right: metrics Spearman's rank correlation matrix, the normalized eigenvalues and the MDS visualization of metrics. The middle panel indicates that 2D visualization is enough. The x-axis in the third panel corresponds to the first eigenvalue.

| Data | SIM | KL | sAUC |
|---|---|---|---|
| OSIE EYE | 0.54±0.06 | 4.71±1.44 | 0.76±0.06 |
| OSIE AMT | 0.43±0.03 | 6.37±0.91 | 0.61±0.03 |

Table 2. The comparison of inter-participant visual congruency between eye tracking (EYE) and mouse tracking amazon mechanical turk (AMT) data. Mean and standard deviation are reported. The smoothing parameter for generating the maps is the optimized value reported in [33, 14].

*data compared to eye tracking?"* The inter-participant[1] visual congruency reflects the amount of consistency among participants in viewing the same image. To compute the amount of IPVC, we follow the one-versus-all scheme as in [29]. That is, we take out one participant and compare it against the fixation map of all other participants. This process is repeated for all the observers and for all the images.

The results are summarized in Table 2, where ANOVA analysis shows all the metrics are significantly different between groups ($\rho < 0.001$). As depicted, there is higher visual congruency between the participants of the eye tracking experiments compared to the mouse tracking. In other words, the mouse tracking data shows a higher dispersion between participants. Since the stimuli is the same, this suggests that mouse tracking data is not as accurate as expected for substituting eye tracking.

We furthermore complement the IPVC by conducting another analysis to measure the performance as a function of the number of participants. The number of participants are kept equal to the number of observers in OSIE EYE for OSIE AMT, that is, 15 participants. Due to larger number of participants in OSIE AMT, we make 10 disjoint folds, covering all the mouse participants of OSIE AMT. For each fold, we randomly select $p$ participants from the participants of a fold and evaluate them against all the participants of that fold. The process is repeated 10 times. Similar procedure is employed for OSIE EYE, except that there is only one fold of participants here. It is worth noting that we keep the smoothing factor to the optimum value reported by [33, 14] that produces the maximum performance for the case of all participants for efficiency reasons.

Fig. 3 summarizes the results. Akin to the IPVC experiment, there exists a higher dispersion between the mouse participants compared to the eye tracking participants. The KL and SIM, however, converge to their upper bounds for maximum participants since they are designed to produce 0 and 1 for exactly similar inputs, respectively. The sAUC always shows a significant gap between mouse and eye data.

**Number of participants.** *"How many participants are required in order to replicate eye tracking by following mouse movements?"* To answer the question, we randomly choose participants from the mouse tracking data of OISE AMT, and build a mouse density map to predict the eye fixation density maps. We evaluate different number of participants. The evaluation process is repeated 10 times and the mean performance is reported.

Fig. 4 depicts the results, indicating a significant gap between mouse tracking and eye tracking. Even 90 participants contributing mouse data can not achieve the eye tracking performance of 15 observers. In one hand, this shows there is no need for more than 40 or 50 mouse participants consistent with the number used in SALICON [14]. On the other hand, the result is alarming as the influence of the existing large gap on saliency models is not well-investigated.

**Contextual information and gaze allocation.** *"How is gaze allocated to different regions and does mouse tracking data capture the same contextual information as eye tracking does?"* The gaze allocation on different image parts and their contextual annotations is investigated by measuring the average allocated gaze over all contextual annotations and images. The results are summarized in Fig. 5. Aligned with the numerous studies on attention guiding features, it is not surprising that regions associated with face, motion, and watchable property are more attended.

Fig. 5 also compares the gaze allocation to annotated regions by eye tracking with gaze allocation by mouse track-ing. The results indicate that on average mouse and eye tracking show the same trend ($\rho = 0.94$ and $\rho = 0.94$ for LAB and AMT, respectively). On a finer scale, however, they have different characteristics. For example, eye tracking shows more gaze allocation to faces than motion, while mouse tracking allocates more attention to motion than faces. Furthermore, eye tracking associates more attention to text, emotion, gazed and watchable regions while mouse tracking affiliate more attention to smell, taste, touch, and operability. Some properties are almost identical in grabbing attention by mouse and eye, e.g., sound and touched.

---
[1]We use the term inter-participant instead of inter-observer to signify the role of other recording mediums such as mouse.

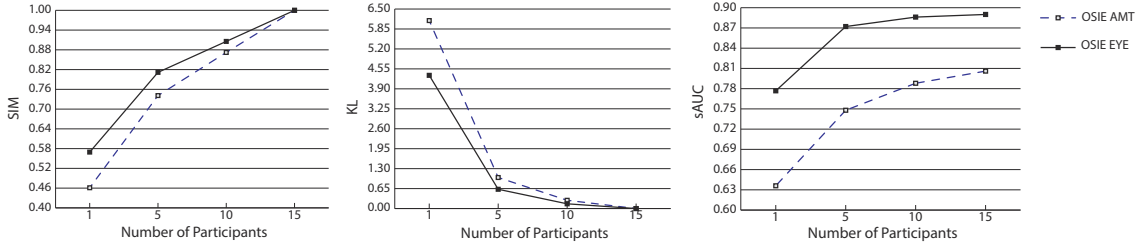

Figure 3. Comparing $p$ participants against all participants. EYE is evaluated by EYE as ground truth and AMT is evaluated using AMT. The smoothing parameter for generating the maps is the optimized value reported in [33, 14].

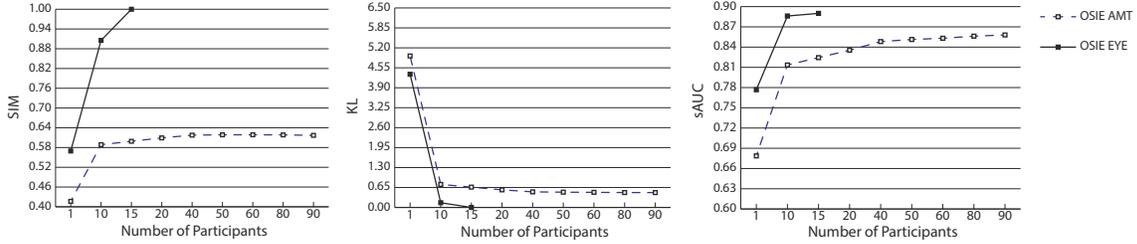

Figure 4. The number of mouse participants to achieve eye tracking performance, i.e., both EYE and AMT are evaluated using EYE as ground truth. The smoothing parameter for generating the maps is the optimized value reported in [33, 14].

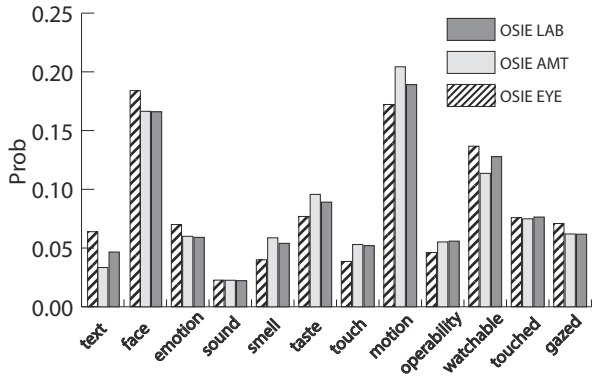

Figure 5. Average allocated gaze on all the contextual annotations using eye tracking and mouse tracking data.

**Contextual performance of mouse maps.** *"How well does mouse tracking capture contextual information in comparison to eye tracking?"* To answer this question, we evaluate mouse density maps against eye tracking ground truth. The contextual evaluation of mouse maps against eye tracking for both OSIE AMT and OSIE LAB is summarized in Fig. 6. Results show a gap between mouse tracking and eye tracking. This gap becomes more significant for some properties, e.g., background regions are the mostly inconsistent areas SIM < 0.6 and KL > 0.8. This can be an indicator that mouse tracking may have a better agreement with eye tracking as long as salient areas are foreground. This finding also agrees with the existence of higher dispersion in mouse tracking, shown by IPVC analysis in Table 2 and experiment of Fig. 3.

**Training on mouse data.** *"What is the effect of training a model by mouse data on its performance?"* We aim to answer this question by training a model on mouse data and evaluating it against eye tracking data in two experiments. In the first experiment, training database is OSIE and the test set is the MIT1003 [18]. We utilize the open source implementation of SALICON model [12], a.k.a OpenSalicon [27]. The training is done using the same initialization and 3 epochs, feeding all the images 3 times using the same random image order, for each ground truth type, including eye tracking (EYE) and mouse tracking data using laboratory (LAB) and Amazon Mechanical Turk (AMT). Since MIT1003 [18] does not have contextual annotation, we only report the traditional scores for it.

Table 3 reports the result of training OpenSalicon using mouse and eye tracking data. The statistical significance test indicates that all the paris of trained OpenSalicon model (by mouse or eye data) are significantly different than each other in terms of the metrics for ($p < 0.001$), except for the sAUC. This indicates that while training using mouse data is similar to training using eye data in terms of sAUC, the generated saliency maps are not necessarily similar to ground truth. Fig. 7 depicts some examples, showing that while the model trained on mouse data captures the salient area, it is prone to over estimation.

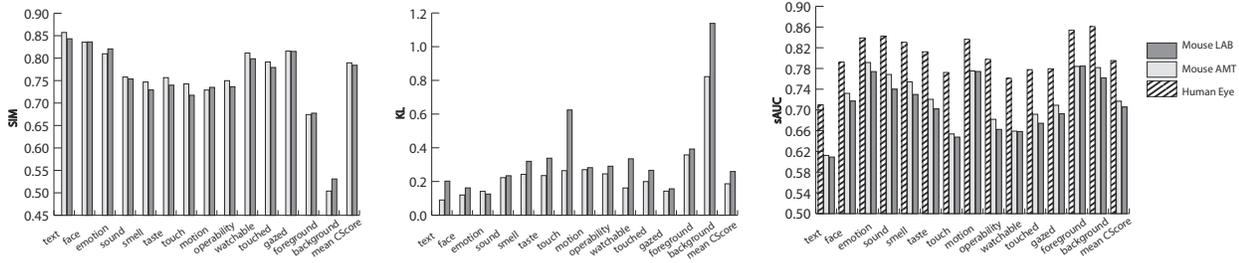

Figure 6. Comparison of mouse density maps and fixation density maps against fixation ground truth (for eye tracking KL=0 and SIM=1).

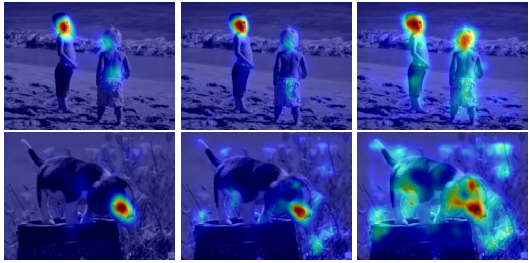

Figure 7. Visual comparison of saliency maps from OpenSalicon, trained by mouse and eye data. From left to right: ground truth, maps from OpenSalicon trained by eye tracking and mouse tracking, respectively.

| Model (Training GT) | SIM | KL | sAUC |
|---|---|---|---|
| Human Performance | 1 | 0 | 0.75 |
| OpenSalicon (EYE) | 0.390 | 1.198 | 0.715 |
| OpenSalicon (AMT) | 0.364 | 1.249 | 0.717 |
| OpenSalicon (LAB) | 0.365 | 1.257 | 0.722 |

Table 3. The effect of training ground truth source on model performance: OpenSalicon performance on MIT1003. The training is carried out on OSIE database using eye tracking (EYE), mouse data from amazon mechanical turk (AMT), and laboratory (LAB) as training ground truth (GT).

The second experiment is carried out on OSIE dataset using a 5 fold cross-validation scheme. That is, one-fifth of the data is used for test and the rest for training. All the images are used in disjoint folds. In each fold, the saliency maps are predicted for the test set and contextually evaluated. The evaluation is always performed using OSIE EYE as ground truth, that is, even if a model is trained on OSIE AMT or LAB, the evaluation of test fold is done using eye tracking-based ground truth.

Figure 8 summarizes the results. Overall, training on eye tracking data achieves a better performance (better mean CScore). There are, however, some differences in terms of contextual performance. For example, training OpenSalicon on mouse tracking data produces better scores for background, while training on eye tracking results in superior scores for foreground. Smell is learned better on mouse tracking AMT ground truth, while faces, emotion, and sound are better captured by a model trained on eye tracking data. To summarize, the findings suggest that the mouse tracking can generally be an acceptable replacement for training data, though the models trained on it can be slightly inferior to the model's trained on eye tracking.

**Evaluation on mouse data.** *"How does evaluation on mouse tracking data affect our understanding of a model's performance?"* We have already observed that there are some differences between mouse tracking and eye tracking in terms of contextual behavior and participant visual congruency. While we could not find severe differences on training a model using either eye tracking or mouse tracking, a crucial question is: will the same phenomenon be observed for model selection and evaluation by mouse data?

To answer this question, we evaluate several models, including, OpenSalicon [27], SalNet [23] (deep network), BMS [35], AWS [10], GBVS [11], Judd [18], eDN [31], CovSal [9] on two databases and ground truth data. The models are chosen based on their performance report on MIT300 [17] and code availability at the time of this writing. We train OpenSalicon on MIT1003 [18]. The pre-trained models of SalNet, eDN, and Judd models, provided by the authors, are used. GBVS, CovSal, AWS, and BMS models do not need training.

The test databases are OSIE, eye tracking and mouse tracking (AMT) ground truths and the MIT300 [17]. We report the traditional sAUC, KL, and SIM scores for each model and database for the sake of comparability between databases. The average rank score of each model, RAS, is computed by averaging over the rank of each score of the model. The models are ranked based on the RAS value.

The results are summarized in Table 4. It shows there is inconsistency between models' ranks on different databases and settings. To investigate the severity of this phenomenon, we computed the Spearman's rank correlation between the pairs of models' ranks on databases. The result reveals that the pair of OSIE eye tracking and MIT300 eye tracking has $\rho = 0.95$ while the pair of OSIE eye tracking and OSIE mouse tracking has the $\rho = 0.73$. Similarly, the pair of OSIE mouse tracking and MIT300 eye tracking has $\rho = 0.80$. To conclude, evaluating models on the same images with different ground truth produces much different

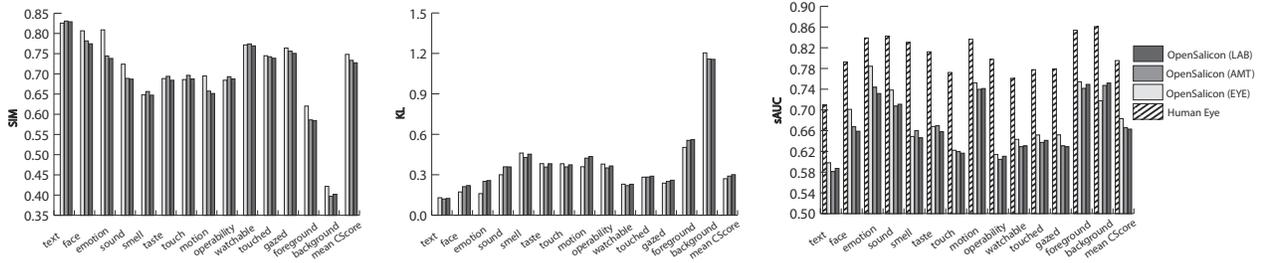

Figure 8. The effect of training ground truth source on model performance. OpenSalicon is trained and tested on OSIE images using 5 fold cross-validation. OpenSalicon (EYE) is trained on eye fixations where OpenSalicon (AMT) and OpenSalicon (LAB) are trained on mouse tracking from amazon mechanical turk and laboratory, respectively. The test ground truth is always eye tracking and the human performance from eye tracking (Human Eye) is reported as the upper bound (for human KL=0 and SIM=1).

ranking compared to evaluating models on different images using eye tracking based ground truth.

This finding adversely affects the reliability of mouse tracking ground truth as an alternative to eye tracking for model evaluation because the rank correlation of saliency models on the same images using different ground truth sources (eye tracking and mouse tracking) is significantly lower than the rank correlation of models by different images and eye tracking ground truth.

## 6. Contextual model evaluation

Over still images, the previous research [8] has shown that models are still consistently under-predicting semantically important image regions (e.g. text, people, actions, and etc.). Computing scores over the whole image and averaging large image collections conceals such deficiencies. Thus, as models continue to improve, measuring more precisely how they preform over contextually annotated regions becomes a necessity. We, here, employ the proposed contextual evaluation scheme for model assessment. Using the models, OpenSalicon [27] (trained on MIT1003), SalNet [23] (deep network), BMS [35], AWS [10], GBVS [11], Judd [18], eDN [31], CovSal [9], we compute the saliency maps for the OSIE EYE and contextually evaluate them.

Table 5 summarizes the results in terms of contextual scores and the mean CScore. Considering the overall score, the mean CScore is numerically different than the traditional scores. While OpenSalicon is the overall top performing model, it is not the winner on all contextual properties. Different models tend to favor different properties. This helps identify a model's deficiency and choose an appropriate model for a specific application. For example, the AWS model is better capturing text areas, while it is not the best model among the current models in terms of mean CScore. Therefore, for a text processing application, AWS may be a better model in order to curtail excess data. More importantly, the proposed scheme allows one to identify the weak points of a saliency model more easily and efficiently.

In terms of model ranking and benchmark, computing the ranks as before, the ranks would be OpenSalicon> SalNet> BMS> AWS> GBVS> eDN> Judd> CovSal. The proposed approach is thus producing similar ranks to the traditional metrics, where the correlation between the model ranks by traditional metrics and mean CScore is $\rho = 0.98$ on OSIE EYE and $\rho = 0.91$ in comparison with the model ranks of MIT300.

## 7. Conclusions and future research

The results of this study show that the inter-participant visual congruency is significantly lower for mouse tracking data in comparison to the eye tracking. We also learn that even 90 mouse tracking participants can not be as accurate as the maps from 15 eye tracking participants. This signifies the inefficiency of the mouse tracking and the fact that less accurate ground truth are obtained by employing mouse tracking. On a fine-grained analysis, this is evident in the disagreement between mouse and eye tracking on background regions. Nonetheless, mouse tracking captures an acceptable level of visual saliency as a low cost alternative to eye tracking.

Analyzing the effect of data type on training deep models using OpenSalicon [27], the open source implementation of [12], reveals that OpenSalicon trained on mouse tracking achieves a sAUC score close to the same model trained on eye tracking data. Further, it captures most salient regions, though it may not produce similar maps to human fixation maps according to SIM and KL scores. This motivates that mouse tracking data can be useful for model training.

In terms of model evaluation, our results are not in favor of mouse tracking data. Mouse tracking seems a less reliable ground truth for model evaluation and ranking. The rank correlation of models are significantly smaller between OSIE EYE and OSIE AMT (same images; different ground truth source type), compared to OSIE EYE and MIT300 (different images; both with eye tracking ground truth), 0.73 vs. 0.95. Due to this, we do not recommended to compare models solely based on mouse tracking data.

| | OSIE database | | | | | | | | | | MIT300 database | | | | |
|---|---|---|---|---|---|---|---|---|---|---|---|---|---|---|---|
| | Eye tracking | | | | | Mouse tracking (AMT) | | | | | Eye tracking | | | | |
| Model | sAUC | KL | SIM | RAS | R | sAUC | KL | SIM | RAS | R | sAUC | KL | SIM | RAS | R |
| Human | 0.89 | 0 | 1 | – | – | 0.77 | 0 | 1 | – | – | 0.81 | 0 | 1 | – | – |
| OpenSalicon [27] | 0.80 | 0.79 | 0.52 | 1 | 1 | 0.68 | 0.55 | 0.67 | 3 | 4 | 0.72 | 0.83 | 0.50 | 2 | 2 |
| SalNet [23] | 0.78 | 0.86 | 0.50 | 2 | 2 | 0.69 | 0.44 | 0.73 | 1 | 1 | 0.69 | 0.81 | 0.52 | 1.33 | 1 |
| BMS [35] | 0.78 | 1.04 | 0.44 | 2.66 | 3 | 0.68 | 0.47 | 0.65 | 2.66 | 3 | 0.65 | 0.81 | 0.51 | 2.33 | 3 |
| AWS [10] | 0.76 | 1.10 | 0.43 | 4 | 4 | 0.68 | 0.47 | 0.64 | 3 | 4 | 0.68 | 1.07 | 0.43 | 4.33 | 5 |
| GBVS [11] | 0.68 | 1.10 | 0.43 | 4.33 | 5 | 0.60 | 0.44 | 0.66 | 2.33 | 2 | 0.63 | 0.87 | 0.48 | 4 | 4 |
| Judd [18] | 0.68 | 1.30 | 0.36 | 5 | 6 | 0.60 | 0.51 | 0.60 | 4 | 6 | 0.60 | 1.12 | 0.42 | 6.33 | 6 |
| eDN [31] | 0.68 | 1.31 | 0.36 | 5.33 | 7 | 0.59 | 0.52 | 0.59 | 5 | 7 | 0.62 | 1.14 | 0.41 | 6.66 | 7 |
| CovSal [9] | 0.59 | 2.26 | 0.40 | 6 | 8 | 0.53 | 2.82 | 0.49 | 6.3 | 8 | 0.57 | 2.68 | 0.47 | 6.66 | 7 |

Table 4. Comparing eye tracking and mouse tracking ground truth data for model evaluation. RAS is the average <u>r</u>ank <u>s</u>core over metrics and R is the final rank.

| Metric | Model | text | face | emotion | sound | smell | taste | touch | motion | operability | watchable | touched | gazed | foreground | background | mean CScore |
|---|---|---|---|---|---|---|---|---|---|---|---|---|---|---|---|---|
| | Human | 0.71 | 0.79 | 0.84 | 0.84 | 0.83 | 0.81 | 0.77 | 0.84 | 0.80 | 0.76 | 0.78 | 0.78 | 0.85 | 0.86 | 0.80 |
| sAUC | OpenSalicon | **0.59** | **0.65** | **0.73** | **0.69** | <u>0.64</u> | **0.65** | <u>0.61</u> | **0.74** | **0.60** | **0.63** | **0.64** | **0.63** | **0.74** | 0.71 | **0.66** |
| | SalNet | 0.55 | **0.65** | <u>0.69</u> | 0.66 | **0.69** | <u>0.64</u> | **0.62** | <u>0.70</u> | **0.60** | <u>0.60</u> | <u>0.60</u> | <u>0.62</u> | 0.68 | <u>0.72</u> | <u>0.64</u> |
| | BMS | 0.56 | <u>0.63</u> | 0.68 | 0.64 | <u>0.64</u> | 0.63 | <u>0.61</u> | 0.63 | <u>0.58</u> | <u>0.60</u> | <u>0.60</u> | <u>0.62</u> | 0.65 | **0.74** | 0.62 |
| | AWS | <u>0.57</u> | 0.60 | 0.60 | 0.62 | 0.61 | 0.60 | 0.60 | 0.63 | 0.57 | 0.59 | 0.59 | 0.60 | 0.64 | **0.74** | 0.60 |
| | eDN | 0.51 | 0.61 | 0.65 | 0.62 | 0.57 | 0.54 | 0.57 | 0.58 | 0.52 | 0.52 | 0.54 | 0.56 | 0.56 | 0.67 | 0.57 |
| | Judd | 0.48 | 0.61 | 0.64 | 0.62 | 0.56 | 0.53 | 0.54 | 0.59 | 0.51 | 0.50 | 0.54 | 0.56 | 0.55 | 0.67 | 0.56 |
| | GBVS | 0.49 | 0.60 | 0.61 | 0.58 | 0.55 | 0.54 | 0.56 | 0.56 | 0.52 | 0.50 | 0.55 | 0.56 | 0.54 | 0.66 | 0.55 |
| | CovSal | 0.47 | 0.61 | 0.63 | 0.59 | 0.52 | 0.49 | 0.52 | 0.52 | 0.50 | 0.48 | 0.51 | 0.56 | 0.50 | 0.61 | 0.53 |
| KL | OpenSalicon | <u>0.13</u> | 0.24 | **0.25** | <u>0.38</u> | 0.46 | <u>0.39</u> | 0.39 | **0.41** | 0.39 | <u>0.24</u> | <u>0.29</u> | **0.27** | 0.55 | <u>1.31</u> | **0.30** |
| | SalNet | 0.30 | <u>0.23</u> | <u>0.29</u> | **0.37** | **0.32** | **0.32** | <u>0.37</u> | **0.46** | **0.33** | 0.33 | **0.28** | 0.30 | <u>0.60</u> | **1.24** | <u>0.32</u> |
| | BMS | <u>0.13</u> | 0.24 | 0.33 | 0.44 | <u>0.43</u> | <u>0.39</u> | 0.39 | 0.62 | 0.38 | **0.23** | 0.32 | **0.27** | 0.74 | 1.41 | 0.35 |
| | GBVS | 0.14 | **0.22** | 0.32 | 0.42 | 0.47 | 0.40 | **0.35** | 0.62 | <u>0.36</u> | 0.26 | 0.32 | **0.27** | 0.78 | 1.34 | 0.35 |
| | AWS | **0.12** | 0.27 | 0.38 | 0.47 | 0.48 | 0.42 | 0.38 | 0.63 | 0.40 | <u>0.24</u> | 0.32 | <u>0.28</u> | 0.78 | 1.40 | 0.37 |
| | eDN | <u>0.13</u> | 0.25 | 0.36 | 0.45 | 0.49 | 0.41 | 0.38 | 0.65 | 0.37 | 0.26 | 0.33 | 0.29 | 0.79 | 1.58 | 0.37 |
| | Judd | 0.14 | 0.26 | 0.36 | 0.44 | 0.49 | 0.42 | 0.39 | 0.65 | 0.38 | 0.26 | 0.34 | 0.29 | 0.80 | 1.59 | 0.37 |
| | CovSal | 2.17 | 1.16 | 0.46 | 0.85 | 1.11 | 1.78 | 2.33 | 1.34 | 1.58 | 2.23 | 0.88 | 1.29 | 1.82 | 3.32 | 1.45 |
| SIM | SalNet | **0.83** | **0.78** | <u>0.72</u> | **0.69** | **0.70** | **0.71** | **0.72** | <u>0.65</u> | **0.72** | **0.78** | **0.75** | **0.76** | <u>0.58</u> | **0.42** | **0.74** |
| | OpenSalicon | <u>0.82</u> | <u>0.77</u> | **0.75** | 0.68 | 0.64 | <u>0.68</u> | 0.68 | **0.67** | 0.68 | <u>0.76</u> | <u>0.74</u> | <u>0.75</u> | **0.60** | <u>0.39</u> | <u>0.73</u> |
| | BMS | <u>0.82</u> | 0.76 | 0.70 | 0.65 | <u>0.65</u> | 0.67 | 0.68 | 0.58 | 0.68 | <u>0.76</u> | 0.72 | 0.74 | 0.54 | 0.35 | 0.71 |
| | GBVS | <u>0.82</u> | <u>0.77</u> | 0.70 | 0.66 | 0.63 | 0.67 | <u>0.70</u> | 0.59 | <u>0.69</u> | 0.75 | 0.72 | 0.74 | 0.51 | 0.37 | 0.71 |
| | AWS | **0.83** | 0.75 | 0.68 | 0.65 | 0.63 | 0.66 | 0.69 | 0.59 | 0.68 | <u>0.76</u> | 0.72 | 0.74 | 0.52 | 0.35 | 0.70 |
| | eDN | <u>0.82</u> | 0.75 | 0.68 | 0.64 | 0.62 | 0.66 | 0.68 | 0.57 | 0.68 | 0.75 | 0.71 | 0.73 | 0.50 | 0.31 | 0.69 |
| | Judd | <u>0.82</u> | 0.75 | 0.68 | 0.65 | 0.62 | 0.65 | 0.68 | 0.57 | 0.67 | 0.75 | 0.71 | 0.73 | 0.50 | 0.30 | 0.69 |
| | CovSal | 0.71 | 0.75 | <u>0.72</u> | 0.66 | 0.58 | 0.59 | 0.62 | 0.55 | 0.63 | 0.66 | 0.68 | 0.70 | 0.44 | 0.38 | 0.66 |

Table 5. Contextual evaluation of models on OSIE with eye tracking as ground truth. The sAUC, KL, and SIM scores for each of the contextual properties, background, foreground and mean CScore, i.e., mean gaze weighted contextual score, are reported. The human performance for KL=0 and SIM=1. Models appear in descending performance order for each score.

**Future directions.** Our results show that mouse tracking data in general offers a first order approximation to eye tracking. The mouse tracking data is useful for model training. A fine-grained analysis, however, highlights the shortcomings of mouse tracking data, in particular the effect of contextual cues such as gaze direction, action end point, etc. (see Fig. 1.) Our results suggest that high performance achieved by recent saliency models, based on deep learning, might be merely due to high volume of training data. Mouse data, although noisy, has been very helpful but that does not necessarily mean that collecting even more mouse data will eventually get us to human level accuracy over fixations.

We believe that future research should focus on fine-grained analysis of ground truth data and models in order to understand attentional mechanism better and improve existing saliency models. Our research showed that there is no single model performing best on all the contextual annotations. This indicates models may be complementary to each other and motivates further research towards understanding models' behaviors on fine-grained details.

**Acknowledgments.** The Finnish Centre of Excellence in Computational Inference Research (COIN) is acknowledged.